\documentclass[a4paper,fleqn,10pt,twocolumn]{AROB-ISBC-SWARM24}

\usepackage{amsmath}
\usepackage{cite}
\usepackage{comment} 
\usepackage{amssymb}

\usepackage[hang,small,bf]{caption}
\usepackage[subrefformat=parens]{subcaption}
\title{DEQ-MCL: Discrete-Event Queue-based Monte-Carlo Localization}

\author{Akira Taniguchi${}^{1\dagger}$, Ayako Fukawa${}^{2}$ and Hiroshi Yamakawa${}^{2,3}$}
\speaker{Akira Taniguchi}

\affils{${}^{1}$Ritsumeikan Univervsity, Shiga, Japan\\
(E-mail: a.taniguchi@em.ci.ritsumei.ac.jp)\\
${}^{2}$The Whole Brain Architecture Initiative, Tokyo, Japan\\
${}^{3}$The University of Tokyo, Tokyo, Japan\\
}
\abstract{%
Spatial cognition in hippocampal formation is posited to play a crucial role in the development of self-localization techniques for robots.
In this paper, we propose a self-localization approach, DEQ-MCL, based on the discrete event queue hypothesis associated with phase precession within the hippocampal formation. Our method effectively estimates the posterior distribution of states, encompassing both past, present, and future states that are organized as a queue. This approach enables the smoothing of the posterior distribution of past states using current observations and the weighting of the joint distribution by considering the feasibility of future states. Our findings indicate that the proposed method holds promise for augmenting self-localization performance in indoor environments.
}


\begin{document}

\maketitle

\section{Introduction}
\label{sec:intro}

Insights from hippocampal formation (HF), an organ in the brain responsible for spatial cognition, may be beneficial for developing robotic self-localization and navigation methods.
HF involves regions such as the hippocampus, dentate gyrus and entorhinal cortex, where functions related to self-localization, such as place and grid cells, have been identified~\cite{Tolman1948a, Okeefe1978placecells}.
Techniques for self-localization and map generation based on hippocampal insights have been exemplified by RatSLAM~\cite{milford2004ratslam}.
Additionally, a phenomenon known as theta phase precession, a characterized behavior of HF, has been discovered.
Theta phase precession involves alterations in the timing of activity in hippocampal neurons according to the theta-wave phase, which compresses the past, present, and future states.
In relation to this, a paper on the probabilistic generative model in HF proposed the discrete-event queue hypothesis~\cite{Taniguchi2021hpf-pgm}.
However, only a hypothesis was presented, and no empirical verification was conducted by implementing a specific computational model.
In this study, we focus on the relationship between this discrete-event queue hypothesis and self-localization.

Generally, in robotics, during navigation, path planning is performed to reach the goal point. Self-localization is performed along the path while moving towards the goal.
Self-localization is essential in path navigation; however, the planned path (future action sequence) is often not effectively used for localization.
If the future action sequence is known in advance, it can be used for self-localization.
Therefore, in this study, we assume that the future action sequence is predetermined by path planning.
Specifically, from the current belief distribution of one's position, after acting in the future, the probability of positions that would hit walls or obstacles in the predicted belief distribution is reduced, thus feeding back the prediction of future states into the current belief distribution.
In other words, it means anticipating: ``If I were in this position now, I might run into a wall, so I probably wouldn't be in such a position.''

In this study, we implement an extended self-localization method using the Bayesian filter and the discrete-event queue hypothesis. Specifically, we represent the joint distribution of states within a fixed time interval, including past, present, and future states.
This extended joint distribution of state variables can be solved using a particle filter, similar to Monte-Carlo localization (MCL)~\cite{dellaert1999monte}.
By incorporating the future action sequence for prediction, the robot estimates the joint distribution while considering the feasibility of the predicted trajectory in the future.


\section{Discrete Event Queue Hypothesis in Hippocampal Formation}
\label{sec:hypothesis}

\begin{figure*}[tb]
    \centering
        \includegraphics[width=0.76\linewidth]{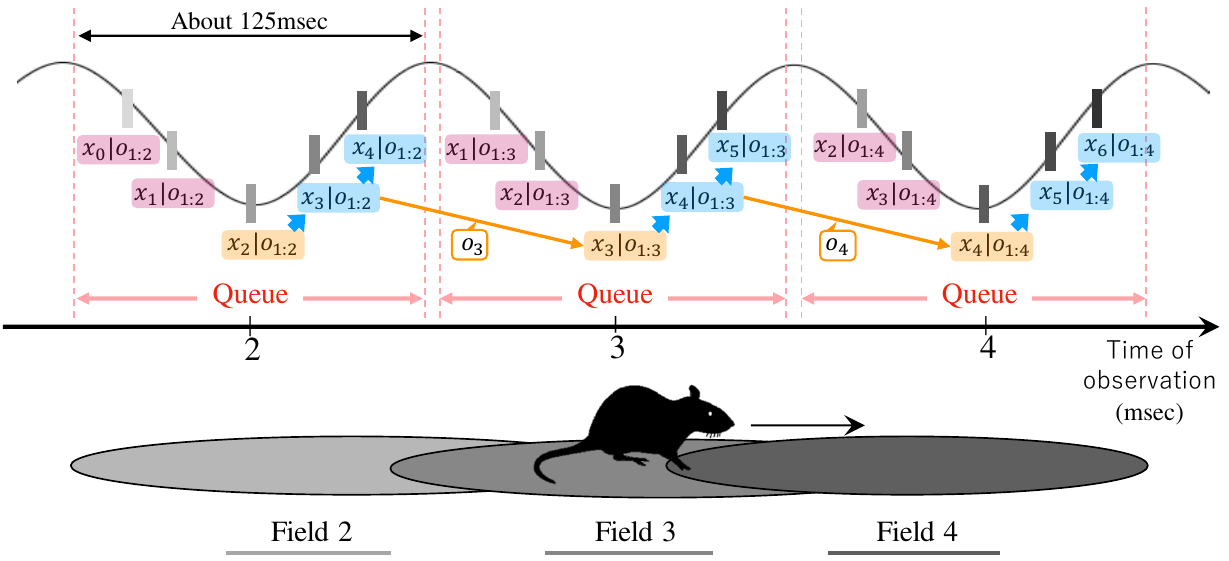}
        \caption{
        Phase precession in theta waves and queue representation.
        The theta phase precession observed in the HF of rodents represents spatial information as a rat moves from left to right.
        Each field indicates the place receptive field that the rat passes through at that time, and the firing timing of place cells shifts from a late phase to an early phase along with the rat's movement.
        In the discrete event queue hypothesis, a queue is assigned for each cycle of the theta wave.
        The variable definition is the same as shown in Fig.~\ref{fig:gm_mcl}.
        The variable \(x_{3} | o_{1:3}\) indicates the estimated state value at time \(t=3\) based on observations up to time \(t=3\).
        Note that the action \(a_{t}\) is omitted from the notation.
        }
        \label{fig:isousaisa}
\end{figure*}

\subsection{Theta Phase Precession in hippocampal Formation}
\label{sec:hpf-pgm:queue}

In HF, as shown in Fig.~\ref{fig:isousaisa}, theta phase precession is considered to discretize experiences and represent the moving trajectory in a compressed form within time steps. External stimuli are sampled in the theta-wave cycles, and current, past, and future events are believed to be encoded in its phase~\cite{Terada2017a}.

\subsection{Abstraction as a Discrete Event Queue}
\label{sec:hpf-pgm:queue:intro}

By abstracting the information within one cycle of the theta wave as a discrete event queue, the queue calculations can be interpreted as an estimation of the current state, correcting past estimations and predicting the future.
To comprehensively model the time process compressed by theta phase precession, the time granularity across the entire model must be elucidated. To avoid complexity, we introduce the hypothesis of a discrete event queue, which views the signal held by theta phase precession as one that includes past, current, and future events at the present time.
From a neuroscience perspective, the analysis of large-scale brain circuits in macaques suggests that HF can be modeled as a discrete event queue with finite buffer capacity~\cite{Misic2014}.
From an engineering perspective, for intelligent systems dealing with hidden Markovian properties, the function of a discrete event queue in retaining the entirety of observational signals in a short time is deemed useful.
%

\section{Proposed method: DEQ-MCL}
\label{sec:proposed}

In this study, we propose a probabilistic self-localization method called the discrete event queue-based Monte-Carlo localization (DEQ-MCL) that is based on the discrete event queue hypothesis.

\subsection{Graphical Model and Generation Process}

\begin{figure}[tb]
    \centering
    \includegraphics[width=0.78\linewidth]{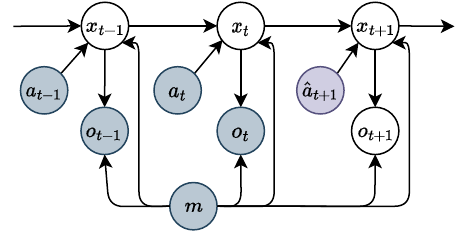}
    \\
        \small
        \begin{tabular}{cl} \hline
            \textit{$m$} & Environmental map \\ 
            \textit{$x_t$} & Robot's state (self-position and angle) \\ 
            \textit{$o_t$} & Observation (depth sensor values) \\ 
            \textit{$a_t$} & Action (motor control values) \\ 
            \textit{$\hat{a}_t$} & Future action (planned control values) \\ %
            \hline
        \end{tabular}
    \caption{
    Graphical model representation and variable definitions for self-localization. 
    White nodes represent latent variables, while colored nodes represent the observed variables or given parameters.
    }
    \label{fig:gm_mcl}
\end{figure}

Our proposed method assumes a graphical model based on the partially observable Markov decision process. 
The graphical model and variable definitions are shown in Fig.~\ref{fig:gm_mcl}.

To capture the influence of the presence or absence of obstacles in the map as a positional probability, we adopt the map-based motion model~\cite{thrun2005probabilistic} as follows:
\begin{align}
    p(x_{t} \mid x_{t-1}, a_{t}, m) \approx p(x_{t} \mid x_{t-1}, a_{t})p(x_{t} \mid m)
    \label{eq:map_motion_model}
\end{align}
In this case, \(p(x_{t} \mid m)\) is a probability distribution based on the configuration space (C-space) or cost map representing the traversability of position \(x_{t}\).

\subsection{Discrete Event Queue Processing}
\label{sec:hpf-pgm:queue:formulation}

\begin{figure}[!tb]
    \centering
    \includegraphics[width=1.00\linewidth]{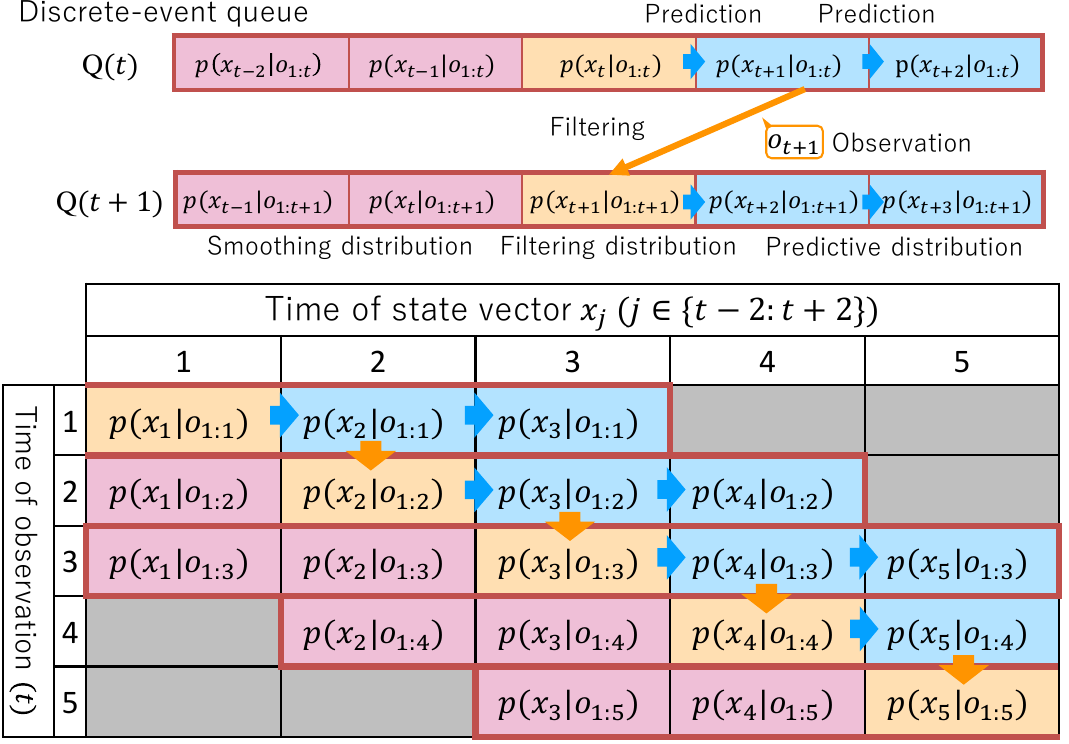}
    \caption{
    Processing of the discrete event queue. In this example, the lag interval of the queue is set to \(L=2\).
    The area enclosed by the thick red border represents the queue that holds the belief distribution of states from \(t-2\) to \(t+2\) at the current time \(t\).
    The horizontal axis of the table at the bottom represents the time of the estimated state, whereas the vertical axis represents the time of observation (corresponding to the horizontal axis of Fig.~\ref{fig:isousaisa}).
    Note that the action \(a_{t}\) is omitted.
    }
    \label{fig:hpf_queue}
\end{figure}

From the hypothesis presented in Section~\ref{sec:hypothesis}, the information retained in phase precession can be interpreted as a discrete event queue.
Theoretically, the discrete event queue can be viewed as a sequential estimation problem of joint posterior distributions over multiple states.
The posterior distribution of the state based on the discrete event queue can be expressed as follows:
\( Q(t) = p(x_{t-L:t+L} \mid a_{1:t}, \hat{a}_{t+1:T}, o_{1:t}, m) \),
where \( L \) is the lag value indicating how far back and forward in time should be included in the queue.
In addition, \( T \) represents the time it takes to reach the goal (planning interval during path planning).

When \( L=2 \), the posterior distribution of the state based on the discrete event queue is as follows:
\begin{align}
    Q(t) 
    &= p(x_{t-2:t+2} \mid a_{1:t}, \hat{a}_{t+1:T}, o_{1:t}, m) \\
    &\propto \underbrace{p(x_{t+2} \mid x_{t+1}, \hat{a}_{t+2})}_{\text{Prediction}} \underbrace{p(x_{t+2} \mid m)}_{\substack{\text{Map-based}\\\text{future state prob.}}} \underbrace{p(o_{t} \mid x_{t}, m)}_{\substack{\text{observation}\\\text{likelihood}}} \nonumber \\
    &\quad \cdot \underbrace{p(x_{t-2:t+1} \mid a_{1:t-1}, \hat{a}_{t:T}, o_{1:t-1}, m)}_{\int Q(t-1) \, dx_{t-3}}, \quad t \geqq 3.
    \label{eq:deq_mcl}
\end{align}
Here, we assume that \( Q(t-1) \) is obtained at the previous time \( t-1 \).
\( \int Q(t-1) dx_{t-3} \) represents the marginalization of \( x_{t-3} \) from \( Q(t-1) \).
In this case, the process shown in Fig.~\ref{fig:hpf_queue} is performed.
Each element of the table represents the conditional marginal distribution of state \( x_{j} \) under the condition of observations up to time \( j \).

From the above equation, \( Q(t) \) can be recursively determined using \( Q(t-1) \).
Equation (\ref{eq:deq_mcl}) can be solved as an extension of MCL using a particle filter.
The estimation of the posterior distribution \( Q(t) \) is conducted as follows:
(i) As a proposal distribution, based on the queue from \( Q(t-1) \) excluding the earliest state \( x_{t-3} \), sample the future position \( x_{t+2} \) from the predictive distribution and add it to the queue for each particle.
(ii) The probability of the current observation \( o_{t} \) and the prior probability of the position based on the map at time \( t+2 \) are used to weight the particles.
(iii) Resample the particles of the queue.
This considers the positional probability based on the map when performing any future action \( \hat{a}_{t+2} \) determined by planning to influence the current position's particles.

Moreover, since the discrete event queue also holds the estimation results of past positions \( x_{t-2:t-1} \) as a joint distribution, the past estimated values within the queue are corrected by the influence of new information, i.e., the current observation likelihood.
In other words, the proposed method also encompasses the estimation of the smoothing distribution \( p(x_{t-2:t-1} \mid a_{1:t-1}, \hat{a}_{t:T}, o_{1:t}, m) \) by fixed-lag smoothing~\cite{kitagawa2014computational}.
This smoothing operation could be interpreted as post-diction.
The discrete event queue process can be represented as an algorithm that combines filtering, smoothing, and prediction, as shown in Fig.~\ref{fig:hpf_queue}.

The computation of this queue is not limited to the probabilistic generative model in Fig.~\ref{fig:gm_mcl}; it can be applied to any probabilistic generative model of a partially observable Markov decision process.

\section{Experiments}
\label{sec:experiment}
In this experiment, the proposed method is implemented on a mobile robot and operated in a simulator environment to evaluate the self-localization performance.

\subsection{Conditions}
The environment covers an area of \(350 \times 300\) pixels, surrounded by outer walls and an inner wall that encloses a rectangle (see Fig.~\ref{fig:result_mcl}).
The robot's initial position is set at the bottom center, and a path is set to complete a single revolution around the environment in a counterclockwise direction.
The mobile robot is equipped with a forward-facing distance sensor and can perform both forward movement and rotation-based actions. The map of the environment is assumed to be known.
The lag value of the queue is set to \(L=20\). 
\(p(x_{t} \mid m) = \exp(-\beta C)\), where \(C\) is the number of times the predicted particle collides with an obstacle.
$\beta$ is the parameter used to determine the degree of influence of C.
Here, we set $\beta=10$.

We compare the localization performance between the proposed DEQ-MCL method and the traditional self-localization methods: MCL, MCL with smoother, and MCL with map-based motion model.
A total of 10 trials is performed for each method.

As an evaluation metric, we used the measured root mean squared error (RMSE) of the true and estimated self-positions.
While performing localization, we recorded the estimation error for each time step.
Here, $x_{t} = ( {\bf x}_{t}, {\bf y}_{t}, {\bf \cos{\theta_{t}}}, {\bf \sin{\theta_{t}}})$.
The estimation error in the localization was evaluated as shown below.
\begin{equation}
    e_{t} = \sqrt{
    \begin{aligned}
    & (\bar{{\bf x}}_{t} - {\bf x}_{t}^{*})^2 + (\bar{{\bf y}}_{t} - {\bf y}_{t}^{*})^2 \\
    & \phantom{=} + ({\cos{\bar{\theta}}}_{t} - {\cos{\theta}}_{t}^{*})^2 + ({\sin{\bar{\theta}}}_{t} - {\sin{\theta}}_{t}^{*})^2 \\
    \end{aligned}
    }
    \label{eq:et1}
\end{equation}
where ${\bf x}_{t}^{*}, {\bf y}_{t}^{*}$, and ${\theta}_{t}^{*}$ denote the true position coordinates and angle of the robot as obtained from the simulator, and 
$\bar{{\bf x}}_{t}$, $\bar{{\bf y}}_{t}$, and $\bar{\theta_{t}}$ represent the mean values of localization coordinates around the particles at time {$t$}. 
After running the localization, we calculated the average of $e_{t}$.

It also measures the entropy and variance of the estimated particle distribution.
These indices indicate the degree of uncertainty and variability of the distribution.
Each indicator is averaged over all trials and time steps.

\subsection{Results}

Table~\ref{tab:result_localization} shows the self-location estimation error and the entropy and variance of the particle distribution.
SD represents the standard deviation of RMSE in 10 trials.
DEQ-MCL can effectively reduce the error, uncertainty, and variance of the distribution compared with conventional MCL methods.

\begin{figure}[tb]
  \begin{minipage}[b]{0.48\linewidth}
    \centering
    \includegraphics[width=\linewidth]{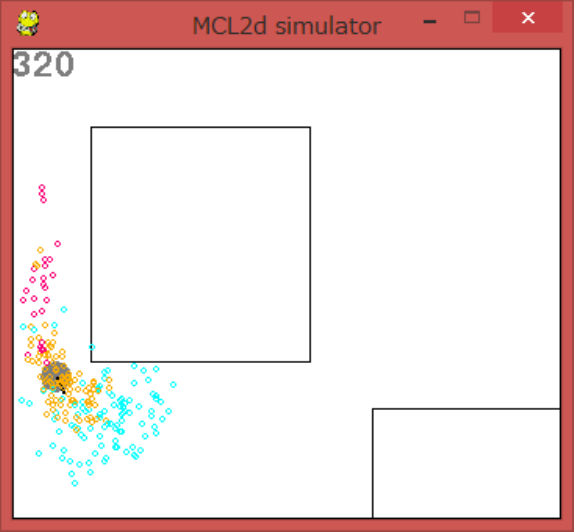}
    \subcaption{DEQ-MCL}
  \end{minipage}
  \begin{minipage}[b]{0.48\linewidth}
    \centering
    \includegraphics[width=\linewidth]{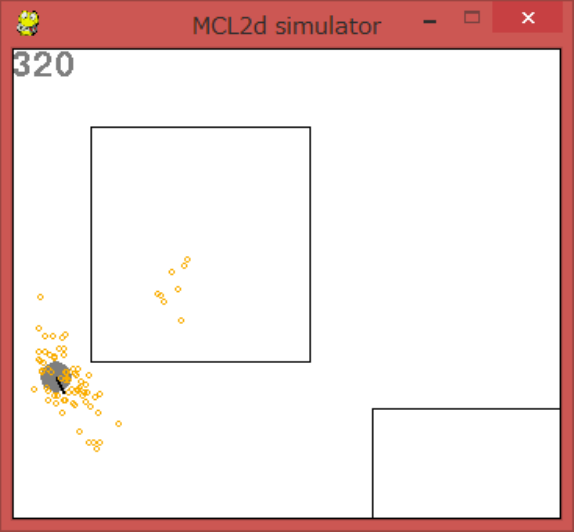}
    \subcaption{MCL}
  \end{minipage}
    \caption{
    Example of self-localization in (a) DEQ-MCL and (b) MCL.
    The black lines represent obstacles, the gray circle represents the robot's position, the orange dots represent the particles of the current state, the pink dots represent the particles of the past state, and the light blue dots represent the particles of the predicted future state. The number in the top left indicates the time step.
    }
    \label{fig:result_mcl}
\end{figure}

An example of self-localization is shown in Fig.~\ref{fig:result_mcl}.
DEQ-MCL not only estimates particles for the current state but also simultaneously estimates past and future states.
Predictions increase uncertainty, leading to a larger spread of particles. However, particles from past states at time \(t-L\) are smoothed, reducing their variance.
Furthermore, compared with MCL, DEQ-MCL demonstrates improvement in reducing misestimations of particles passing through walls, and the spread of particles is better controlled.

\begin{figure}[tb]
  \begin{minipage}[b]{0.48\linewidth}
    \centering
    \includegraphics[width=\linewidth]{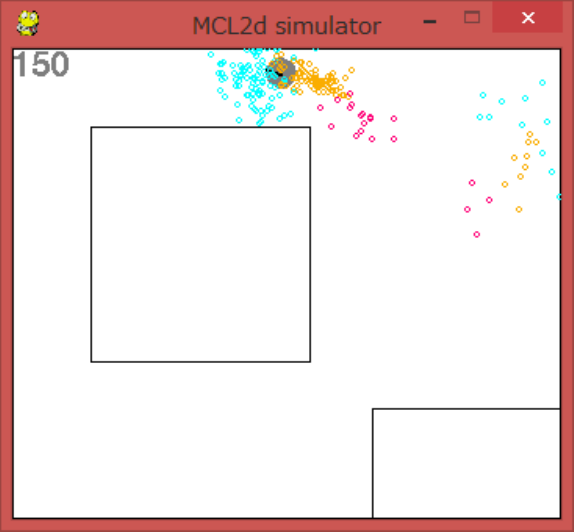}
    \subcaption{DEQ-MCL (step 150)}
  \end{minipage}
  \begin{minipage}[b]{0.48\linewidth}
    \centering
    \includegraphics[width=\linewidth]{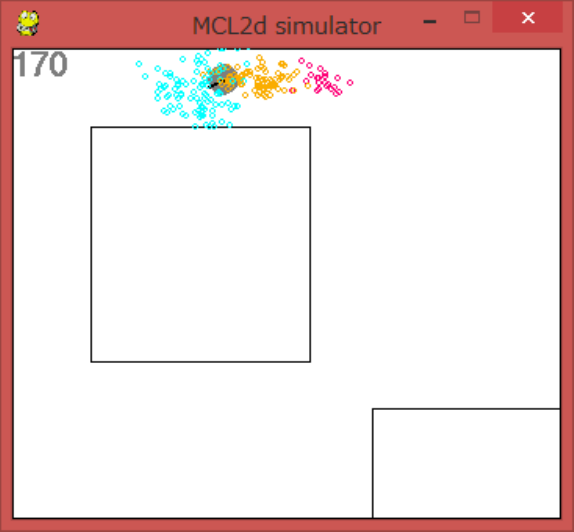}
    \subcaption{DEQ-MCL (step 170)}
  \end{minipage}
    \caption{
    Comparison of particles at present at step 150 (orange in (a)) and particles in the past at step 170 (pink in (b)).
    The pink particles represent the estimated distribution 20 steps ago.
    }
    \label{fig:result_deq_mcl}
\end{figure}

Figure~\ref{fig:result_deq_mcl} shows the DEQ-MCL estimation results with a 20-step difference. The orange particles in Fig.~\ref{fig:result_deq_mcl}(a) and the pink particles in Fig.~\ref{fig:result_deq_mcl}(b) show the self-localization results for the same time step. Notably, a misestimated particle can be observed in the upper right corner of Fig.~\ref{fig:result_deq_mcl}(a). However, Fig.~\ref{fig:result_deq_mcl}(b) illustrates that the post-diction of the self-position effectively eliminates this mis-estimation.

\begin{table*}[tb]
\centering
\caption{Experimental results: Performance of self-localization.}
\small
\begin{tabular}{lccc}
\hline
\textbf{Methods} & \textbf{RMSE (SD)$\downarrow$} & \textbf{Distribution Entoropy$\downarrow$} & \textbf{Distribution Variance$\downarrow$} \\
\hline
DEQ-MCL & \underline{\textbf{20.95 (17.66)}} & \underline{\textbf{7.82}} & 
\begin{tabular}{@{}c@{}}$\bf{x}$: \underline{\textbf{419.92}},  \(\cos \theta\): \underline{\textbf{0.09}} \\ $\bf{y}$: \underline{\textbf{290.04}}, \(\sin \theta\): \underline{\textbf{0.07}}\\ \end{tabular}
\\
\hline
MCL with smoother & \underline{22.88 (20.01)} & \underline{8.88} & 
\begin{tabular}{@{}c@{}}$\bf{x}$:977.39, \(\cos \theta\): \underline{0.14} \\ $\bf{y}$:611.06, \(\sin \theta\): \underline{0.11}\\ \end{tabular}
\\
\hline
MCL with map-based motion model & 25.00 (18.38) & 9.69 & 
\begin{tabular}{@{}c@{}}$\bf{x}$:\underline{792.28}, \(\cos \theta\): 0.17 \\ $\bf{y}$:616.85, \(\sin \theta\): 0.16 \\ \end{tabular}
\\
\hline
MCL & 26.17 (22.00) & 9.81 & 
\begin{tabular}{@{}c@{}}$\bf{x}$:966.93, \(\cos \theta\): 0.19 \\ $\bf{y}$:\underline{591.79}, \(\sin \theta\): 0.16 \\ \end{tabular}
\\
\hline
\end{tabular}
\label{tab:result_localization}
\end{table*}

\section{Conclusion}
\label{sec:conclusion}

Inspired by phase precession in the HF, a brain region responsible for spatial cognition, we proposed a self-localization method based on the discrete event queue hypothesis that includes past, present, and future states.
The proposed method improved self-localization performance by using observations up to the present to correct past state estimates and feedback by incorporating future predictions into current state estimates.

This study also provides insights into the functional aspect of self-localization in the phase precession of HF.
In the future, we intend to investigate the consistency of such functionality with neuroscientific findings to determine if it is genuinely realized in the brain.
While this study focused on the self-localization function of HF, we also intend to discuss the relationship between methods related to memory consolidation and learning~\cite{ataniguchi2020spcoslam2}, as well as active inference~\cite{friston2016} and control~\cite{Stahl2011} in the future.
%


\section*{Acknowledgements}
This research was partially supported by JSPS Grants-in-Aid JP22H05159 and JP23K16975.




\bibliographystyle{IEEEtran}
\bibliography{DEQ-MCL}


\end{document}